\documentclass[final]{cvpr}

\usepackage{times}
\usepackage{epsfig}
\usepackage{graphicx}
\usepackage{amsmath}
\usepackage{amssymb}

\usepackage{xcolor}
\usepackage{multirow}
\usepackage{amssymb}
\usepackage{amsmath}
\usepackage{colortbl}
\usepackage{threeparttable}
\usepackage[switch]{lineno}

\usepackage[pagebackref=false,breaklinks=true,colorlinks,bookmarks=false]{hyperref}

\def\cvprPaperID{7869} 
\def\confYear{CVPR 2021}
\begin{document}

\title{Camouflaged Object Segmentation with Distraction Mining}

\author{
Haiyang Mei\textsuperscript{1} \ \ \ \ \ \ 
Ge-Peng Ji\textsuperscript{2,4} \ \ \ \ \ \  
Ziqi Wei\textsuperscript{3,$\star$} \ \ \ \ \ \  
Xin Yang\textsuperscript{1,$\star$} \ \ \ \ \ \ 
Xiaopeng Wei\textsuperscript{1} \ \ \ \ \ \ 
Deng-Ping Fan\textsuperscript{4}
\\
\textsuperscript{1} Dalian University of Technology \ \ \ \ \ \ 
\textsuperscript{2} Wuhan University \ \ \ \ \ \ 
\textsuperscript{3} Tsinghua University \ \ \ \ \ \   
\textsuperscript{4} IIAI
\\
{\tt\small \url{https://mhaiyang.github.io/CVPR2021_PFNet/index}}
}

\maketitle

\renewcommand{\thefootnote}{}

\footnotetext{\textsuperscript{$\star$} Xin Yang (xinyang@dlut.edu.cn) and Ziqi Wei are the corresponding authors.}

\thispagestyle{empty}

\begin{abstract}
Camouflaged object segmentation (COS) aims to identify objects that are ``perfectly'' assimilate into their surroundings, which has a wide range of valuable applications.
The key challenge of COS is that there exist high intrinsic similarities between the candidate objects and noise background.
In this paper, we strive to embrace challenges towards effective and efficient COS.
To this end, we develop a bio-inspired framework, termed \textbf{P}ositioning and \textbf{F}ocus \textbf{Net}work (\textbf{PFNet}), which mimics the process of predation in nature.
Specifically, our PFNet contains two key modules, i.e., the positioning module (PM) and the focus module (FM).
The PM is designed to mimic the detection process in predation for positioning the potential target objects from a global perspective and the FM is then used to perform the identification process in predation for progressively refining the coarse prediction via focusing on the ambiguous regions.
Notably, in the FM, we develop a novel \textbf{distraction mining strategy} for the distraction discovery and removal, to benefit the performance of estimation.
Extensive experiments demonstrate that our PFNet runs in real-time (72 FPS) and significantly outperforms 18 cutting-edge models on three challenging datasets under four standard metrics.
\end{abstract}

\section{Introduction}

Camouflage is the concealment of animals or objects by any combination of material, coloration, or illumination, for making the target objects hard to see (crypsis) or disguising them as something else (mimesis) \cite{stevens2009animal}. Benefiting from the capability of finding out the camouflaged objects that are ``seamlessly'' embedded in their surroundings, camouflaged object segmentation (COS) has a wide range of valuable applications in different fields, ranging from medical diagnosis (\emph{e.g.}, polyp segmentation~\cite{fan2020pra_pranet} and lung infection segmentation~\cite{fan2020inf}), industry (\emph{e.g.}, inspection of unqualified products on the automatic production line), agriculture (\emph{e.g.}, locust detection to prevent invasion), security and surveillance (\emph{e.g.}, search-and-rescue mission and the detection of pedestrians or obstacles in bad weather for automatic driving), scientific research (\emph{e.g.}, rare species discovery), to art (\emph{e.g.}, photo-realistic blending and recreational art).

\begin{figure}[tbp]
	\begin{center}
		\includegraphics[width=1\linewidth, height=3.6cm]{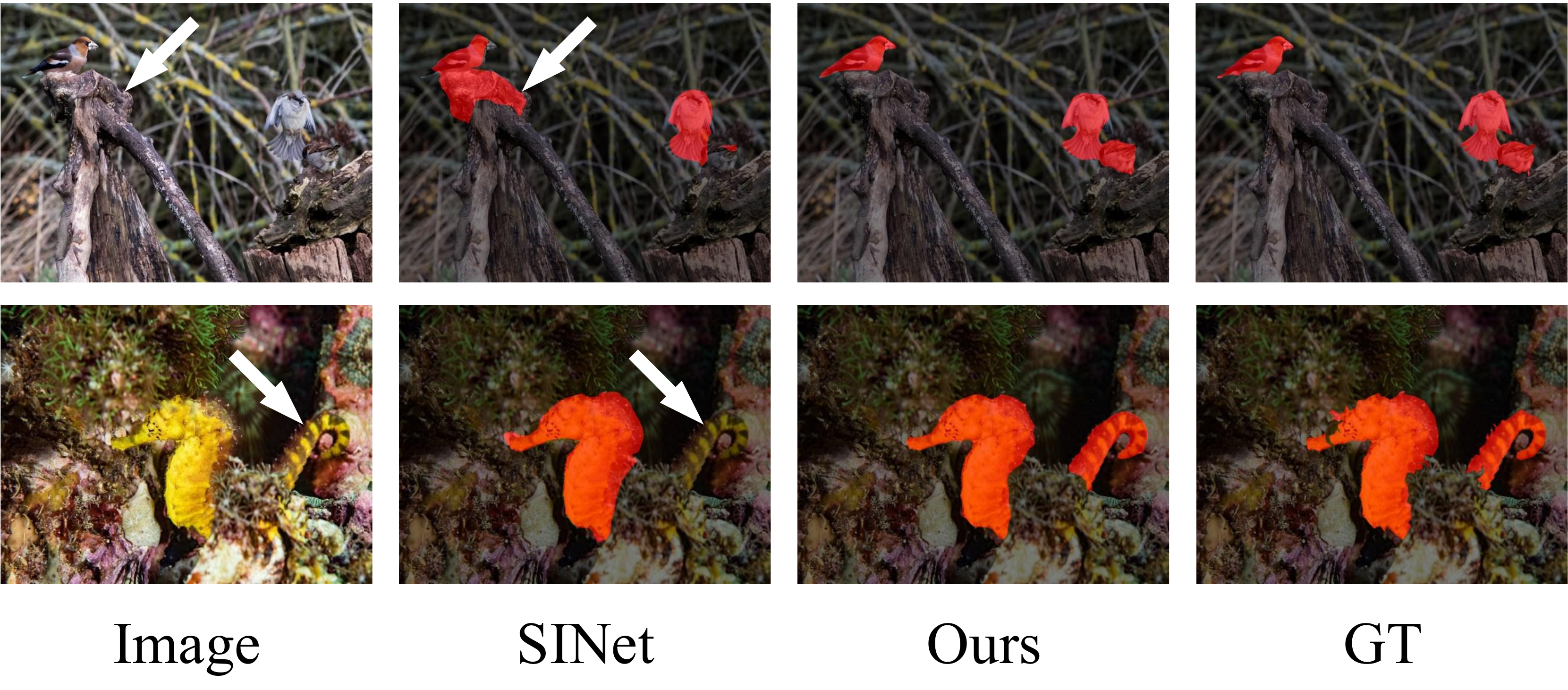}
	\end{center}
	\caption{Visual examples of camouflaged object segmentation. While the state-of-the-art method SINet \cite{Fan_2020_CVPR_sinet} confused by the background region which shares similar appearance with the camouflaged objects (pointed to by a arrow in the top row) or the camouflaged region that cluttered in the background (pointed to by a arrow in the bottom row), our method can eliminate these distractions and generate accurate segmentation results.}
	\label{fig:teaser}
\end{figure}

However, COS is a fundamentally challenging task due to the fact that the camouflage strategy works by deceiving the visual perceptual system of the observer \cite{stevens2009animal} and thus a significant amount of visual perception knowledge \cite{troscianko2009camouflage} is required to eliminate the ambiguities caused by the high intrinsic similarities between the target object and the background.
Research into camouflaged object segmentation has a long and rich history in many fields such as biology and art \cite{stevens2009animal}. Early methods are dedicated to distinguishing the foreground and background based on handcrafted low-level features such as texture \cite{4579856}, 3D convexity \cite{pan2011study} and motion \cite{li2011detection}. These features, however, have limited capability to distinguish between the camouflaged and non-camouflaged objects, so the approaches based on them often fail in complex scenes. Despite the recently proposed deep learning-based approaches~\cite{le2019anabranch_camo,Fan_2020_CVPR_sinet,yan2020mirrornet} have achieved performance improvement to some extent, there is still a large room for exploring the effective way of accurate COS.

In nature, prey animals make use of mechanisms such as camouflage to misdirect the visual sensory mechanisms of predators for reducing the risk of being detected \cite{stevens2009animal}. Under the pressure of natural selection, predatory animals have evolved a variety of adaptations such as sharp senses and intelligent brains for the successful predation which can be divided into three stages, \emph{i.e.}, detection, identification, and capture \cite{hall2013camouflage}. This motivates our bio-inspired solution to segment camouflaged objects by mimicking the first two stages of predation.

In this paper, we propose a positioning and focus network (PFNet) which greatly improves the existing camouflaged object segmentation performance. Our PFNet contains two key modules, \emph{i.e.}, the positioning module (PM) and the focus module (FM). The PM is designed to mimic the detection process in predation for positioning the potential target objects from a global perspective and the FM is then used to perform the identification process in predation for refining the initial segmentation results by focusing on the ambiguous regions.
Specifically, the PM consists of a channel attention block and a spatial attention block and both of them are implemented in a non-local way to capture long-range semantic dependencies in terms of channel and spatial position for inferring the initial location of the target objects from a global perspective.
The FM first perform multi-scale context exploration based on the foreground-attentive (background-attentive) features for discovering the false-positive (false-negative) distractions and then remove these distractions to get the purer representations about the target objects. Such distraction mining strategy is implemented in an implicit way and is applied on different levels of features to progressively refine the segmentation results, enabling our PFNet to possess the strong capability of accurately segmenting the camouflaged objects (see Figure \ref{fig:teaser} as an example).
To sum up, our contributions are as follows:
\begin{itemize}
	\item We introduce the concept of distraction to the COS problem and develop a novel distraction mining strategy for distraction discovery and removal, to benefit the accurate segmentation of the camouflaged object.
	\item We propose a new COS framework, named positioning and focus network (PFNet), which first positioning the potential target objects by exploring long-range semantic dependencies and then focuses on distraction discovery and removal to progressively refine the segmentation results.
	\item We achieve state-of-the-art camouflaged object segmentation performance on three benchmark datasets. Experimental results demonstrate the effectiveness of our method.
\end{itemize}

\section{Related Work}

\textbf{Generic Object Detection (GOD)} seeks to locate object instances from several predefined generic categories in natural images \cite{Liu2018Deep_god}, which is one of the most fundamental and challenging problems in computer vision and forms the basis for solving complex or high-level vision tasks such as segmentation \cite{kirillov2019panoptic}, scene understanding \cite{li2009towards_total}, and object tracking \cite{yilmaz2006object_survey}. The generic objects in a scene can be either conspicuous or camouflaged, and the camouflaged ones can be seen as hard cases. Therefore, directly applying GOD methods (\emph{e.g.}, \cite{lin2017feature_fpn,he2017mask_maskrcnn,huang2019mask_msrcnn}) to segment camouflaged objects may not get the desired results.

\textbf{Salient Object Detection (SOD)} aims to identify and segment the most attention-grabbing object(s) in an input image. Hundreds of image-based SOD methods have been proposed in the past decades \cite{Fan_2018_ECCV_clutter}. Early methods are mainly based on the handcrafted low-level features as well as heuristic priors (\emph{e.g.}, color \cite{achanta2009frequency} and contrast \cite{cheng2014global}). Recently, deep convolutional neural networks (CNNs) have set new state-of-the-art on salient object detection. Multi-level feature aggregation is explored for robust detection \cite{Lee_2016_CVPR,HouPami19Dss,Zhang_2017_ICCV,Zhao_2019_CVPR_pfan}. Recurrent and iterative learning strategies are also employed to refine the saliency map progressively \cite{Zhang_2018_CVPR_progressive,Wang_2019_CVPR_iterative}. Due to the effectiveness for feature enhancement, attention mechanisms \cite{vaswani2017attention,Woo_2018_ECCV_cbam} are also applied to saliency detection \cite{Liu_2018_CVPR_picanet,Chen_2018_ECCV_ras}.
In addition, edge/boundary cues are leveraged to refine the saliency map \cite{Qin_2019_CVPR_basnet,zhao2019egnet,Su_2019_ICCV_banet}.
However, applying the above SOD approaches for camouflaged object segmentation may not appropriate as the term ``salient'' is essentially the opposite of ``camouflaged'', \emph{i.e.}, standout versus immersion.

\textbf{Specific Region Segmentation (SRS)} we defined here refers to segmenting the specific region such as shadow \cite{Hu_2018_CVPR_dsc,Le_2018_ECCV_ad_net,Zhu_2018_ECCV_bdrar,zheng2019distraction}, mirror \cite{Yang_2019_ICCV_mirror,Mei_2021_CVPR_PDNet}, 
glass \cite{Mei_2020_CVPR_glass,xie2020segmenting_glass} and water \cite{Han_2018_ECCV_water} region in the scene. Such regions are special and has a critical impact on the vision systems. For the water, shadow and mirror region, there typically exists intensity or content discontinuities between the foreground and background. Instead, both the intensity and content are similar between the camouflaged objects and the background, leading to a great challenge of COS. Besides, the camouflaged objects are typically with more complex structures, compared with the glass region, and thus increasing the difficulty of accurate segmentation.

\begin{figure*}[tbp]
	\begin{center}
		\includegraphics[width = 1\linewidth]{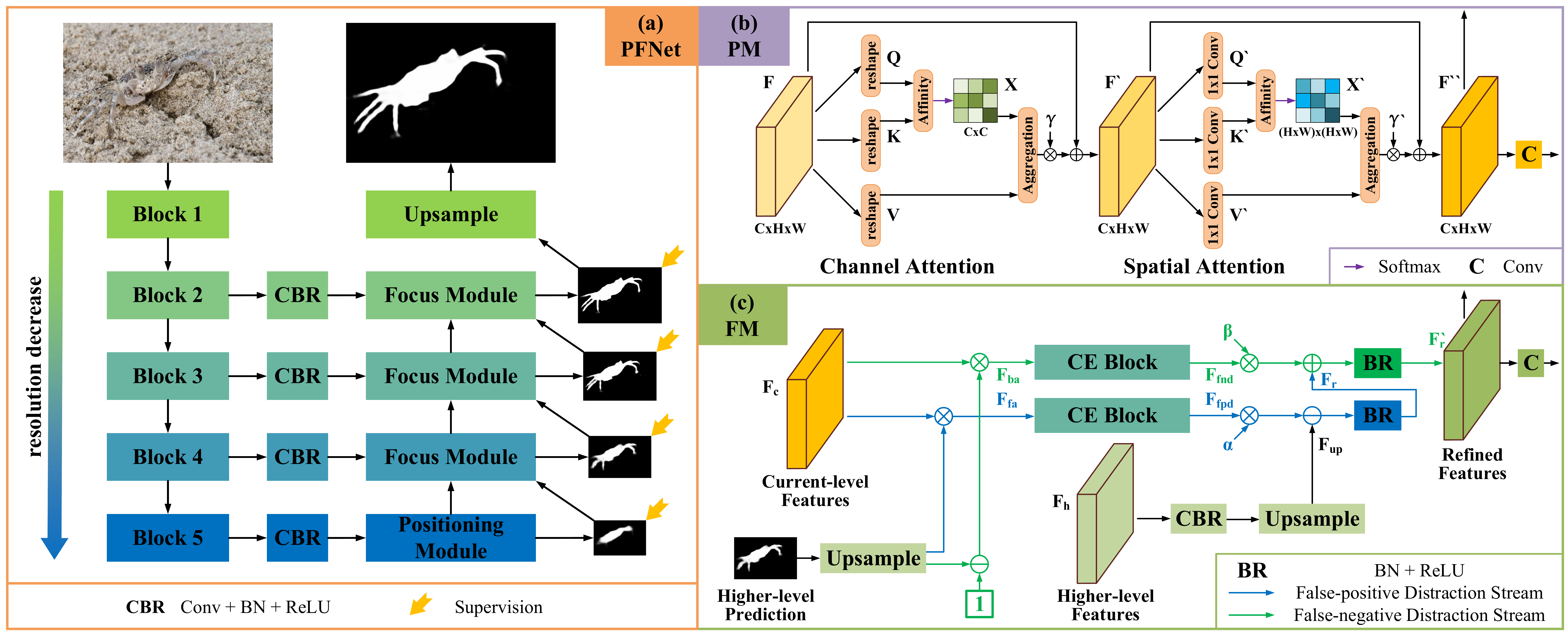}
	\end{center}
	\caption{(a) Overview of our positioning and focus network (PFNet) and its two main building blocks: (b) a positioning module (PM) and (c) a focus module (FM).}
	\label{fig:pipeline}
\end{figure*}

\textbf{Camouflaged Object Segmentation (COS)} has a long and rich research history in many fields such as biology and art \cite{stevens2009animal}, and hugely influenced by two remarkable studies \cite{Thayer1909Concealing,Cott1940}.
Early works related to camouflage are dedicated to distinguishing foreground and background based on the handcrafted low-level features such as texture \cite{4579856}, 3D convexity \cite{pan2011study} and motion \cite{li2011detection}. These methods work for a few simple cases but often fail in complex scenes. Recently, Le \emph{et al.} \cite{le2019anabranch_camo} propose an end-to-end network for camouflaged object segmentation through integrating classification information into pixel-level segmentation. Yan \emph{et al.} \cite{yan2020mirrornet} further introduce the adversarial attack to boost up the segmentation accuracy. Fan \emph{et al.} \cite{Fan_2020_CVPR_sinet} develop a simple yet effective framework, termed as \emph{SINet}, and construct the current largest COS dataset \emph{COD10K} to facilitate the advance of the COS in the deep learning era.

\textbf{Contextual Feature Learning} plays an important role in achieving high performance for many computer vision tasks. Many works are devoted to exploiting contexts to enhance the ability of feature representation. Specifically, multi-scale contexts are developed in \cite{chen2017deeplab,Zhao_2017_CVPR_pspnet,Mei_2021_TCSVT_DCENet} and multi-level contexts are extracted in \cite{yang2019drfn,Zhang_2020_ICME}. Large-field contextual features are captured in \cite{peng2017large,Mei_2020_CVPR_glass}, direction-aware contexts are explored in \cite{Hu_2018_CVPR_dsc}, and contextual contrasted features are leveraged in \cite{Ding_2018_CVPR_contrast,Yang_2019_ICCV_mirror}.
However, exploring contextual features indiscriminately may not contribute much to COS as the contexts would often be dominated by features of conspicuous objects. Our method differs from the above works by focusing on exploring contexts from the foreground/background-attentive features for contextual reasoning and distraction discovery. And we validate the effectiveness of our method by the experiments.

\section{Methodology}
It has been pointed in the biological study \cite{hall2013camouflage} that the process of predation can be broken down into three stages, \emph{i.e.}, detection, identification and capture.
Inspired by the first two stages of predation, we design a positioning and focus network (PFNet) which consists of two key modules, \emph{i.e.}, the positioning module (PM) and the focus module (FM).
The PM is designed to mimic the detection process in predation for positioning the potential target objects from a global perspective and the FM is then used to perform the identification process in predation for refining the initial segmentation results by focusing on the ambiguous regions.

\subsection{Overview}
The overview of our proposed network is shown in Figure \ref{fig:pipeline}~(a). Given a single RGB image, we first feed it into a ResNet-50 \cite{he2016deep_resnet} backbone to extract multi-level features which are further fed into four convolution layers for channel reduction. Then, a positioning module (PM) is applied on the highest-level features to locate the potential target objects. Finally, multiple focus modules (FMs) are leveraged to progressively discover and remove both false-positive and false-negative distractions, for the accurate identification of the camouflaged object.

\subsection{Positioning Module}
Figure \ref{fig:pipeline}~(b) illustrates the detailed structure of the well-designed positioning module (PM). Given the input highest-level features, the PM aims to harvest semantic-enhanced high-level features and further generate the initial segmentation map. It consists of a channel attention block and a spatial attention block. Both of them are implemented in a non-local way, to capture long-range dependencies in terms of channel and spatial position, for enhancing the semantic representation of the highest-level features from a global perspective.

Specifically, given the input features $F \in \mathbb{R}^{C\times H\times W}$, where $C$, $H$, and $W$ represent the channel number, height, and width, respectively, we first reshape $F$ to get the query $Q$, key $K$, and value $V$, respectively, where $\{Q, K, V\}\in\mathbb{R}^{C\times N}$ and $N=H\times W$ is the number of pixels. Then we perform a matrix multiplication between $Q$ and the transpose of $K$, and apply a softmax layer to calculate the channel attention map $X \in \mathbb{R}^{C\times C}$:
\begin{equation}\label{equ:x}
	x_{ij} = \frac{exp(Q_{i:}\cdot K_{j:})}{\sum_{j=1}^{C}exp(Q_{i:}\cdot K_{j:})},
\end{equation}
where $Q_{i:}$ denotes the $i$-$th$ row of matrix $Q$ and $x_{ij}$ measures the $j^{th}$ channel's impact on the $i^{th}$ channel. After that, we perform a matrix multiplication between $X$ and $V$ and reshape the aggregated attentive features to $\mathbb{R}^{C\times H\times W}$. Finally, to enhance the fault-tolerant ability, we multiply the result by a learnable scale parameter $\gamma$ and perform an identify mapping operation to obtain the final output $F'\in\mathbb{R}^{C\times H\times W}$:
\begin{equation}\label{equ:Fc}
	F'_{i:} = \gamma\sum_{j=1}^{C}(x_{ij}V_{j:})+F_{i:},
\end{equation}
where $\gamma$ gradually learns a weight from an initial value of 1. The final feature $F'$ models the long-range semantic dependencies between the channels of feature maps and thus is more discriminative than the input feature $F$.

Then, we feed the output features of channel attention block into the spatial attention block as the input. We first employ three $1\times 1$ convolution layers on the input features $F'$ and reshape the convolution results to generate three new feature maps $Q'$, $K'$, and $V'$, respectively, where $\{Q', K'\}\in\mathbb{R}^{C_1\times N}$ and $C_1=C/8$, and $V'\in\mathbb{R}^{C\times N}$. After that we perform a matrix multiplication between the transpose of $Q'$ and $K'$, and use the softmax normalization to generate the spatial attention map $X'\in\mathbb{R}^{N\times N}$:
\begin{equation}\label{equ:x'}
	x'_{ij} = \frac{exp(Q'_{:i}\cdot K'_{:j})}{\sum_{j=1}^{N}exp(Q'_{:i}\cdot K'_{:j})},
\end{equation}
where $Q'_{:i}$ denotes the $i$-$th$ column of matrix $Q'$ and $x'_{ij}$ measures the $j^{th}$ position's impact on the $i^{th}$ position. Meanwhile, we conduct a matrix multiplication between $V'$ and the transpose of $X'$ and reshape the result to $\mathbb{R}^{C\times H\times W}$. Similar to the channel attention block, we multiply the result by a learnable scale parameter $\gamma'$ and add a skip-connection to obtain the final output $F''\in\mathbb{R}^{C\times H\times W}$:
\begin{equation}\label{equ:Fp}
	F''_{:i} = \gamma'\sum_{j=1}^{N}(V'_{:j}x'_{ji})+F'_{:i},
\end{equation}
where $\gamma'$ is also initialized as 1. Based on $F'$, $F''$ further gains the semantic correlations between all positions and thus enhancing the semantic representation of the feature.

Finally, we can get the initial location map of the targets by applying a $7\times 7$ convolution with the padding of 3 on $F''$. The $F''$ and the initial location map would be refined progressively by the following focus modules (FMs).

\begin{figure}[tbp]
	\begin{center}
		\includegraphics[width = 1\linewidth]{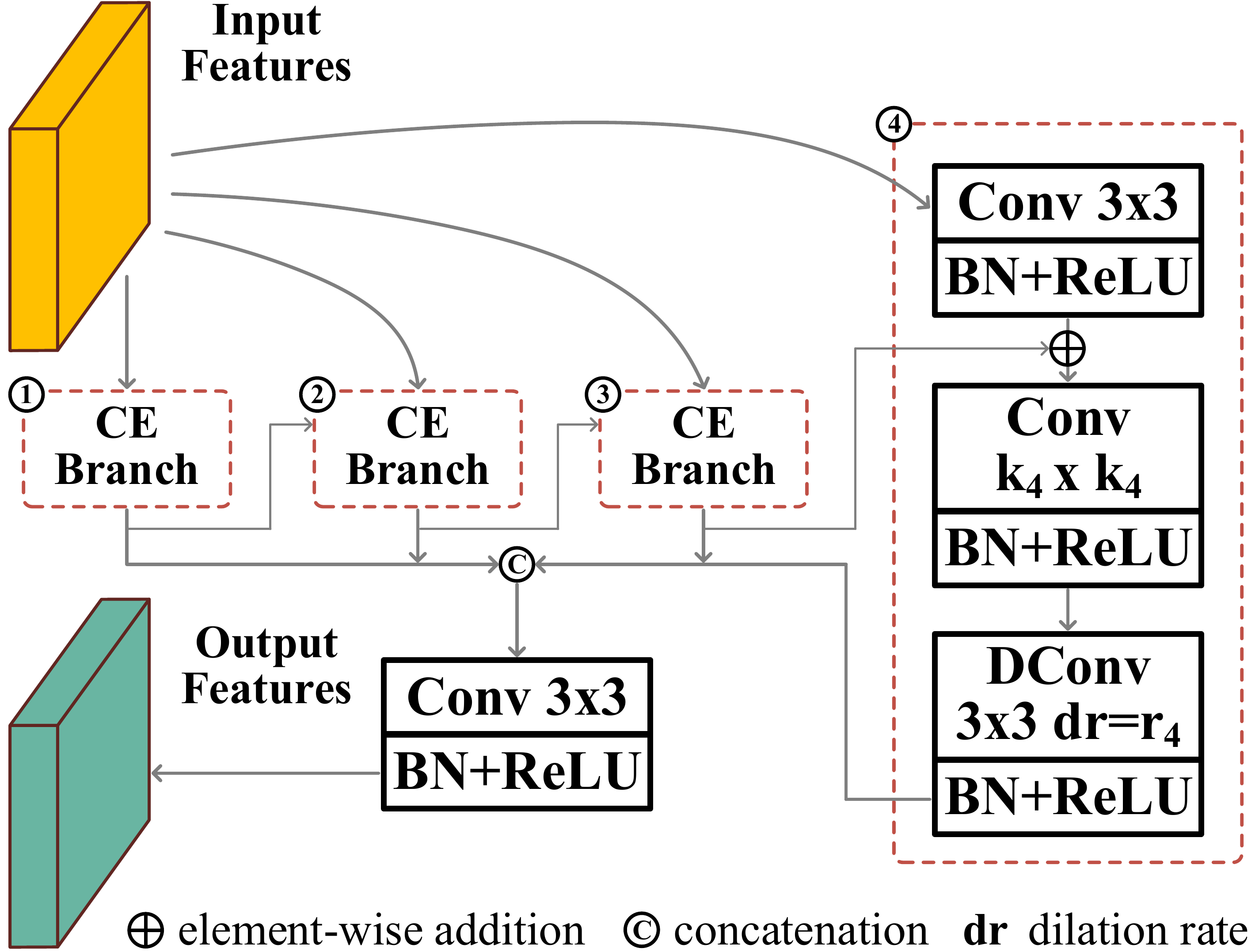}
	\end{center}
	\caption{\small The architecture of our context exploration (CE) block.}
	\label{fig:ceblock}
\end{figure}

\subsection{Focus Module}
As the camouflaged objects typically share a similar appearance with the background, both false-positive and false-negative predictions would naturally occur in the initial segmentation. The focus module (FM) is designed to first discover and then remove these false predictions. It takes the current-level features derived from the backbone and the higher-level prediction and features as the input, and outputs the refined features and a more accurate prediction.

\textbf{Distraction discovery.} We note that humans could distinguish the distractions well after a careful analysis. Our observation is that humans would do context reasoning, \emph{i.e.}, comparing the patterns (\emph{e.g.}, texture and semantic) of the ambiguous regions with that of the confident regions, to make the final decision. This inspires us to conduct contextual exploration for all the predicted foreground (or background) regions, for the purpose of discovering the false-positive distractions (or the false-negative distractions) that are heterogeneous with the confident foreground (or background) prediction regions.
As shown in Figure \ref{fig:pipeline}~(c), we first upsample the higher-level prediction and normalize it with a sigmoid layer. Then we use this normalized map and its reverse version to multiply the current-level features $F_c$, to generate the foreground-attentive features $F_{fa}$ and the background-attentive features $F_{ba}$, respectively. Finally, we feed these two types of features into two parallel context exploration (CE) blocks to perform contextual reasoning for discovering the false-positive distractions $F_{fpd}$ and the false-negative distractions $F_{fnd}$, respectively.

As shown in Figure \ref{fig:ceblock}, the CE block consists of four context exploration branches and each branch includes a $3\times 3$ convolution for channel reduction, a $k_i\times k_i$ convolution for local feature extraction, and a $3\times 3$ dilated convolution with a dilation rate of $r_i$ for context perceiving. We set $k_i, i\in\{1,2,3,4\}$ to 1, 3, 5, 7, and set $r_i, i\in\{1,2,3,4\}$ to 1, 2, 4, 8, respectively. Each convolution is followed by a batch normalization (BN) layer and a ReLU nonlinearity operation. The output of the $i^{th}, i\in\{1,2,3\}$ CE branch will be fed into $(i+1)^{th}$ branch to be further processed in a larger receptive field. The outputs of all four branches are then concatenated and fused via a $3\times 3$ convolution. By such design, the CE block gains the capability of perceiving rich contexts over a wide range of scales and thus could be used for context reasoning and distraction discovery.

\textbf{Distraction Removal.} After distraction discovery, we can perform distraction removal in the following way:
\begin{equation}\label{equ:removal}
	\begin{aligned}
		F_{up} &= U(CBR(F_{h})), \\
		F_{r} &= BR(F_{up} - \alpha F_{fpd}),\\
		F'_{r} &= BR(F_{r} + \beta F_{fnd}),
	\end{aligned}
\end{equation}
where $F_{h}$ and $F'_{r}$ denote the input higher-level features and the output refined features, respectively; CBR is the combination of convolution, batch normalization (BN) and ReLU; U is the bilinear upsampling; and $\alpha$ and $\beta$ are the learnable scale parameters which are initialized as 1. Here we use the element-wise subtraction operation to suppress the ambiguous backgrounds (\emph{i.e.}, false-positive distractions) and the element-wise addition operation to augment the missing foregrounds (\emph{i.e.}, false-negative distractions).

Finally, a more accurate prediction map can be obtained by applying a convolution layer on the refined feature $F'_{r}$. We use the ground truth map to supervise the generated map, to force the $F'_{r}$ into a purer representation than $F_h$, \emph{i.e.}, the distraction removed features. This would further guide the CE block to discover the specific form of distractions and make the whole focus module (FM) works on distraction discovery and removal in an implicit way.
Note that we do not adopt the specific distraction map to explicitly supervise the $F_{fpd}$ and $F_{fnd}$, based on the following two considerations: (\romannumeral1) annotating false positives and false negatives are both expensive and subjective, making it difficult to obtain sufficient and representative distractions; and (\romannumeral2) using a fixed distraction supervision for all focus modules (FMs) is suboptimal as the input higher-level features for each FM is different and the distractions we hope to discover and remove should vary dynamically with the gradually refined input higher-level features.

\textbf{Discussion.} Distraction cues have been explored in many vision tasks such as salient object detection \cite{Chen_2018_ECCV_ras,xiao2018deep_distraction}, semantic segmentation \cite{huang2017semantic} and visual tracking \cite{zhu2018distractor}. Existing works leverage either the false-positive distraction \cite{huang2017semantic,xiao2018deep_distraction,zhu2018distractor} or the false-negative distraction \cite{Chen_2018_ECCV_ras} to obtain more accurate results. Unlike the above methods, we explore both two types of distractions and propose a well-designed focus module (FM) to first discover and then remove these distractions.
Although the distraction-aware shadow (DS) module in \cite{zheng2019distraction} also exploits both two distractions, our proposed focus module (FM) is inherently different from the DS module in the following three aspects. First, DS module extracts features to predict two types of distractions based on the same input features while our focus module (FM) finds false-positive distraction from the foreground-attentive features and discovers false-negative distraction from the background-attentive features. Second, the feature extractor in the DS module contains two $3\times3$ convolutions while our context exploration (CE) block consists of four branches, which could capture multi-scale contexts for better distraction discovery. Third, the supervision for the DS module is acquired  based on the differences between the predictions from existing shadow detection models (\emph{i.e.}, \cite{Hu_2018_CVPR_dsc,Zhu_2018_ECCV_bdrar,Le_2018_ECCV_ad_net}) and the ground truths. Such an explicit supervision strategy would be constrained by the specific methods and thus may have limited generality. By contrast, we design an implicit distraction mining strategy via imposing ground truth supervision on the distraction-removed features to force each CE block exploring the specific form of distractions. To the best of our knowledge, we are the first to mine distractions for camouflaged object segmentation and we believe that the proposed strategy of distraction mining could provide insights to other vision tasks.

\begin{table*}[t]
	\centering
	\footnotesize
	\renewcommand\arraystretch{1.0}
	\setlength{\tabcolsep}{6.0pt}
	\begin{tabular}{l|c|cccc|cccc|cccc}
		\hline
		\multirow{2}{*}{Methods} & \multirow{2}{*}{Pub.'Year} &  \multicolumn{4}{c|}{CHAMELEON (76 images)} & \multicolumn{4}{c|}{CAMO-Test (250 images)} & \multicolumn{4}{c}{COD10K-Test (2,026 images)} \\
		\cline{3-14}
		& & $S_\alpha$$\uparrow$ & $E_\phi^{ad}$$\uparrow$ & $F_\beta^w$$\uparrow$ & $M\downarrow$ & $S_\alpha$$\uparrow$ & $E_\phi^{ad}$$\uparrow$ & $F_\beta^w$$\uparrow$ & $M\downarrow$ & $S_\alpha$$\uparrow$ & $E_\phi^{ad}$$\uparrow$ & $F_\beta^w$$\uparrow$ & $M\downarrow$  \\
		
		\hline
		
		FPN$^\circ$ \cite{lin2017feature_fpn}             & CVPR'17 & 0.794 & 0.835 & 0.590 & 0.075 & 0.684 & 0.791 & 0.483 & 0.131 & 0.697 & 0.711 & 0.411 & 0.075     \\
		PSPNet$^\bullet$ \cite{zhao2017pyramid_pspnet}           & CVPR'17 & 0.773 & 0.814 & 0.555 & 0.085 & 0.663 & 0.778 & 0.455 & 0.139 & 0.678 & 0.688 & 0.377 & 0.080     \\
		Mask RCNN$^\star$ \cite{he2017mask_maskrcnn}        & ICCV'17 & 0.643 & 0.780 & 0.518 & 0.099 & 0.574 & 0.716 & 0.430 & 0.151 & 0.613 & 0.750 & 0.402 & 0.080     \\
		UNet++$^\S$ \cite{zhou2018unet++}          & DLMIA'17 & 0.695 & 0.808 & 0.501 & 0.094 & 0.599 & 0.740 & 0.392 & 0.149 & 0.623 & 0.718 & 0.350 & 0.086     \\
		DSC$^\vartriangle$ \cite{Hu_2018_CVPR_dsc}       & CVPR'18 & 0.850 & 0.888 & 0.714 & 0.050 & 0.736 & 0.830 & 0.592 & 0.105 & 0.758 & 0.788 & 0.542  & 0.052 \\
		PiCANet$^\dagger$ \cite{liu2018picanet}          & CVPR'18 & 0.769 & 0.836 & 0.536 & 0.085 & 0.609 & 0.753 & 0.356 & 0.156 & 0.649 & 0.678 & 0.322 & 0.090     \\
		BDRAR$^\vartriangle$ \cite{Zhu_2018_ECCV_bdrar}    & ECCV'18 & 0.779 & 0.881 & 0.663 & 0.064 & 0.759 & 0.825 & 0.664 & 0.093 & 0.753 & 0.836 & 0.591 & 0.051 \\
		HTC$^\star$ \cite{chen2019hybrid_htc}             & CVPR'19 & 0.517 & 0.490 & 0.204 & 0.129 & 0.476 & 0.442 & 0.174 & 0.172 & 0.548 & 0.521 & 0.221 & 0.088     \\
		MSRCNN$^\star$ \cite{huang2019mask_msrcnn}           & CVPR'19 & 0.637 & 0.688 & 0.443 & 0.091 & 0.617 & 0.670 & 0.454 & 0.133 & 0.641 & 0.708 & 0.419 & 0.073     \\
		BASNet$^\dagger$ \cite{qin2019basnet}           & CVPR'19 & 0.687 & 0.742 & 0.474 & 0.118 & 0.618 & 0.719 & 0.413 & 0.159 & 0.634 & 0.676 & 0.365 & 0.105     \\
		CPD$^\dagger$ \cite{wu2019cascaded_cpd}             & CVPR'19 & 0.853 & 0.878 & 0.706 & 0.052 & 0.726 & 0.802 & 0.550 & 0.115 & 0.747 & 0.763 & 0.508 & 0.059     \\
		PFANet$^\dagger$ \cite{zhao2019pyramid_pfan}           & CVPR'19 & 0.679 & 0.732 & 0.378 & 0.144 & 0.659 & 0.735 & 0.391 & 0.172 & 0.636 & 0.619 & 0.286 & 0.128     \\
		EGNet$^\dagger$ \cite{zhao2019egnet}           & ICCV'19 & 0.848 & 0.879 & 0.702 & 0.050 & 0.732 & 0.827 & 0.583 & 0.104 & 0.737 & 0.777 & 0.509 & 0.056     \\
		F3Net$^\dagger$ \cite{wei2019f3net}    & AAAI'20 & 0.854 & 0.899 & 0.749 & 0.045 & 0.779 & 0.840 & 0.666 & 0.091 & 0.786 & 0.832 & 0.617 & 0.046     \\
		GCPANet$^\dagger$ \cite{chen2020global_gcpanet}    & AAAI'20 & 0.876 & 0.891 & 0.748 & 0.041 & 0.778 & 0.842 & 0.646 & 0.092 & 0.791 & 0.799 & 0.592 & 0.045     \\
		PraNet$^\S$ \cite{fan2020pra_pranet}       & MICCAI'20 & 0.860 & 0.898 & 0.763 & 0.044 & 0.769 & 0.833 & 0.663 & 0.094 & 0.789 & 0.839 & 0.629 & 0.045     \\
		MINet-R$^\dagger$ \cite{Pang_2020_CVPR_minet}    & CVPR'20 & 0.844 & 0.919 & 0.746 & 0.040 & 0.749 & 0.835 & 0.635 & 0.090 & 0.759 & 0.832 & 0.580 & 0.045     \\
		SINet* \cite{Fan_2020_CVPR_sinet}           & CVPR'20 & 0.869 & 0.899 & 0.740 & 0.044 & 0.751 & 0.834 & 0.606 & 0.100 & 0.771 & 0.797 & 0.551 & 0.051     \\
		
		\hline
		\textbf{PFNet}* & Ours
		& \textbf{0.882} & \textbf{0.942} & \textbf{0.810} & \textbf{0.033}
		& \textbf{0.782} & \textbf{0.852} & \textbf{0.695} & \textbf{0.085}
		& \textbf{0.800} & \textbf{0.868} & \textbf{0.660} & \textbf{0.040} \\
		
		\hline
	\end{tabular}
	\caption{Comparison of our proposed method and other 18 state-of-the-art methods in the relevant fields on three benchmark datasets in terms of the structure-measure $S_\alpha$ (larger is better),  the adaptive E-measure $E_\phi^{ad}$ (larger is better), the weighted F-measure $F_\beta^w$ (larger is better), and the mean absolute error $M$ (smaller is better). All the prediction maps are evaluated with the same code. The best results are marked in \textbf{bold}. $\circ$: object detection method. $\bullet$: semantic segmentation method. $\star$: instance segmentation methods. $\vartriangle$: shadow detection methods. $\S$: medical image segmentation methods. $\dagger$: SOD methods. *: COS methods. Our method outperforms other counterparts with a large margin under all four standard evaluation metrics on all three benchmark datasets.}
	\label{tab:comparison}
\end{table*}

\subsection{Loss Function}
There are four output predictions in the PFNet, \emph{i.e.}, one from the positioning module (PM) and three from the focus modules (FMs). For the PM, we impose binary cross-entropy (BCE) loss $\ell_{bce}$ and IoU loss $\ell_{iou}$ \cite{qin2019basnet} on its output, \emph{i.e.}, $\mathcal{L}_{pm} = \ell_{bce} + \ell_{iou}$, to guide the PM to explore the initial location of the target object. For the FM, we hope it could focus more on the distraction region. Such region is typically located at the object's boundaries, elongated areas or holes. Thus we combine the weighted BCE loss $\ell_{wbce}$ \cite{wei2019f3net} and weighted IoU loss $\ell_{wiou}$ \cite{wei2019f3net}, \emph{i.e.}, $\mathcal{L}_{fm} = \ell_{wbce} + \ell_{wiou}$, to force the FM pay more attention to the possible distraction region. Finally, the overall loss function is:
\begin{equation}\label{equ:overall_loss}
	\mathcal{L}_{overall} = \mathcal{L}_{pm} + \sum_{i=2}^{4}2^{(4-i)}\mathcal{L}_{fm}^{i},
\end{equation}
where $\mathcal{L}_{fm}^{i}$ denotes the loss for the prediction of the focus module at $i$-\emph{th} level of the PFNet.

\section{Experiments}
\subsection{Experimental Setup}
\textbf{Datasets.} We evaluate our method on three benchmark datasets: CHAMELEON \cite{skurowski2018animal_chameleon}, CAMO \cite{le2019anabranch_camo}, and COD10K \cite{Fan_2020_CVPR_sinet}. CHAMELEON \cite{skurowski2018animal_chameleon} has 76 images collected from the Internet via the Google search engine using ``camouflaged animal'' as a keyword and corresponding manually annotated object-level ground-truths. CAMO \cite{le2019anabranch_camo} contains 1,250 camouflaged images covering different categories, which are divided into 1,000 training images and 250 testing images. COD10K \cite{Fan_2020_CVPR_sinet} is currently the largest benchmark dataset, which includes 5,066 camouflaged images (3,040 for training and 2,026 for testing) downloaded from multiple photography websites, covering 5 super-classes and 69 sub-classes. We follow previous work \cite{Fan_2020_CVPR_sinet} to use the training set of CAMO \cite{le2019anabranch_camo} and COD10K \cite{Fan_2020_CVPR_sinet} as the training set (4,040 images) and others as testing sets.

\noindent \textbf{Evaluation Metrics.}
We use four widely used and standard metrics to evaluate our method: structure-measure ($S_\alpha$) \cite{fan2017structure_smeasure}, adaptive E-measure ($E_\phi^{ad}$) \cite{fan2018enhanced_emeasure}, weighted F-measure ($F_\beta^w$) \cite{margolin2014evaluate_wfmeasure}, and mean absolute error ($M$).
Structure-measure ($S_\alpha$) focuses on evaluating the structural information of the prediction maps, which is defined as: $S_\alpha = \alpha S_o + (1-\alpha)S_r$, where $S_o$ and $S_r$ denote the object-aware and region-aware structural similarity, respectively; and $\alpha$ is set to be 0.5 as suggested in \cite{fan2017structure_smeasure}.
E-measure ($E_\phi$) simultaneously evaluates the pixel-level matching and image-level statistics, which is shown to be related to human visual perception \cite{fan2018enhanced_emeasure}. Thus, we include this metric to assess the overall and localized accuracy of the camouflaged object segmentation results.
F-measure ($F_\beta$) is a comprehensive measure on both the precision and recall of the prediction map. Recent studies \cite{fan2017structure_smeasure,fan2018enhanced_emeasure} have suggested that the weighted F-measure ($F_\beta^w$) \cite{margolin2014evaluate_wfmeasure} can provide more reliable evaluation results than the traditional $F_\beta$. Thus, we also consider this metric in the comparison.
The mean absolute error ($M$) metric is widely used in foreground-background segmentation tasks, which calculates the element-wise difference between the prediction map and the ground truth mask.

\begin{figure*}[t]
	\begin{center}
		\includegraphics[width = 1\linewidth]{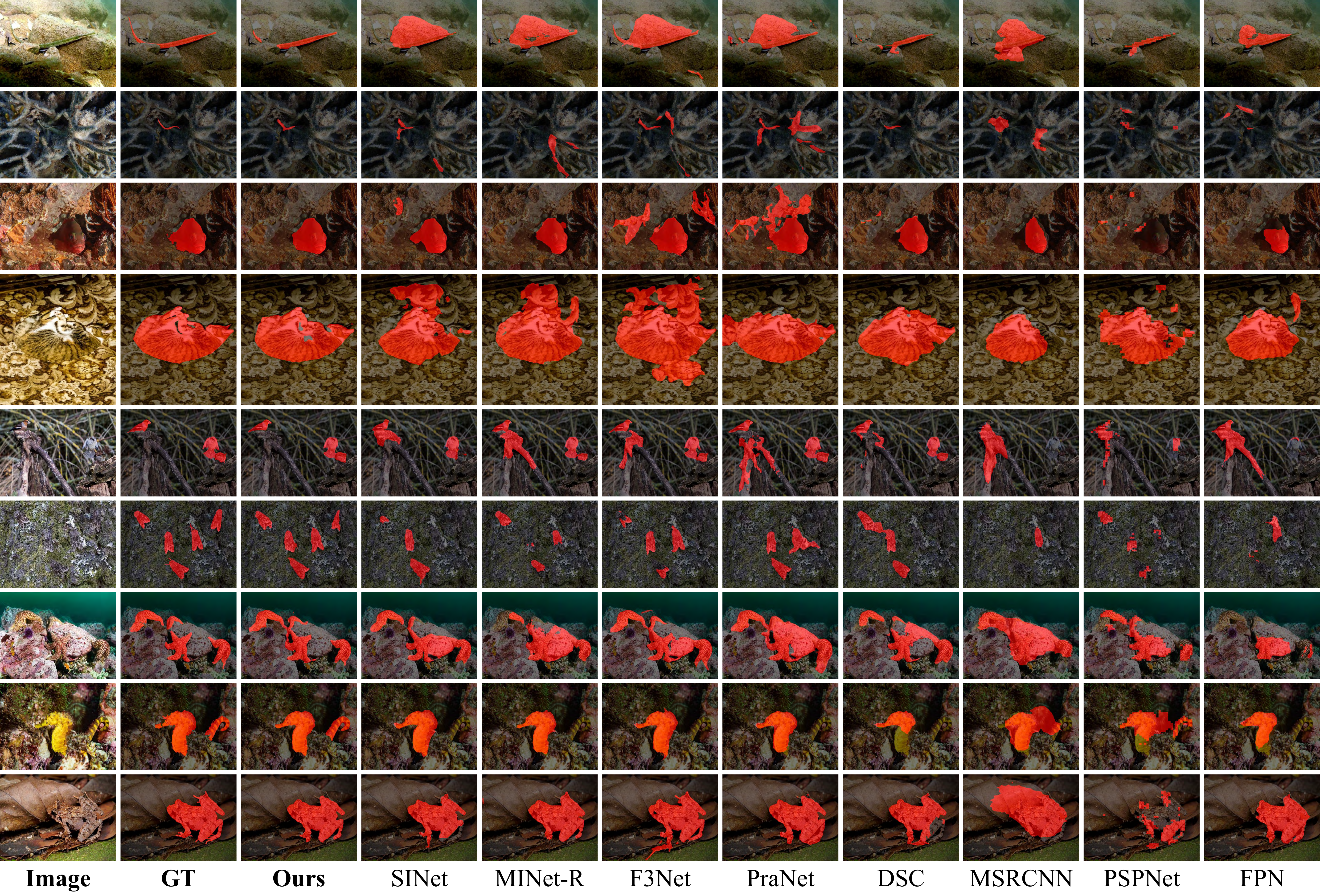}
	\end{center}
	\caption{Visual comparison of the proposed model with state-of-the-art methods. Obviously, our approach is capable of segmenting various camouflaged objects concealed in different environments more accurately.}
	\label{fig:visual}
\end{figure*}

\noindent \textbf{Implementation Details.} We implement our model with the PyTorch toolbox \cite{paszke2019pytorch}. An eight-core PC with an Intel Core i7-9700K 3.6 GHz CPU (with 64GB RAM) and an NVIDIA GeForce RTX 2080Ti GPU (with 11GB memory) is used for both training and testing. For training, input images are resized to a resolution of $416\times416$ and are augmented by randomly horizontal flipping and color jittering. The parameters of the encoder network are initialized with the ResNet-50 model \cite{he2016deep_resnet} pre-trained on ImageNet while the remaining layers of our PFNet are initialized randomly. We use the stochastic gradient descent (SGD) optimizer with the momentum of 0.9 and the weight decay of $5\times10^{-4}$ for loss optimization. We set the batch size to 16 and adjust the learning rate by the poly strategy \cite{liu2015parsenet_poly} with the basic learning rate of 0.001 and the power of 0.9. It takes only about 76 minutes for the network to converge for 45 epochs. For testing, the image is first resized to $416\times416$ for network inference and then the output map is resized back to the original size of the input image. Both the resizing processes use bilinear interpolation. We do not use any post-processing such as the fully connected conditional random field (CRF) \cite{krahenbuhl2011efficient_crf} to further enhance the final output. The inference for a $416\times416$ image takes only 0.014 seconds (about 72 FPS).

\noindent \textbf{Compared Methods.} To demonstrate the effectiveness of our PFNet, we compare it against 18 state-of-the-art baselines: object detection method FPN \cite{lin2017feature_fpn}; semantic segmentation method PSPNet \cite{zhao2017pyramid_pspnet}; instance segmentation methods Mask RCNN \cite{he2017mask_maskrcnn}, HTC \cite{chen2019hybrid_htc}, and MSRCNN \cite{huang2019mask_msrcnn}; shadow detection methods DSC \cite{Hu_2018_CVPR_dsc} and BDRAR \cite{Zhu_2018_ECCV_bdrar}; medical image segmentation methods UNet++ \cite{zhou2018unet++} and PraNet \cite{fan2020pra_pranet}; salient object detection methods PiCANet \cite{liu2018picanet}, BASNet \cite{qin2019basnet}, CPD \cite{wu2019cascaded_cpd}, PFANet \cite{zhao2019pyramid_pfan}, EGNet \cite{zhao2019egnet}, F3Net \cite{wei2019f3net}, GCPANet \cite{chen2020global_gcpanet}, and MINet-R \cite{Pang_2020_CVPR_minet}; and camouflaged object segmentation method SINet \cite{Fan_2020_CVPR_sinet}. For fair comparison, all the prediction maps of the above methods are either provided by the public website or produced by running the models retrained with open source codes. Besides, all the prediction maps are evaluated with the same code.

\subsection{Comparison with the State-of-the-arts}
Table \ref{tab:comparison} reports the quantitative results of PFNet against other 18 state-of-the-art methods on three benchmark datasets. We can see that our method outperforms all the other methods with a large margin on all four standard metrics. For example, compared with the state-of-the-art COS method SINet \cite{Fan_2020_CVPR_sinet}, our method improves $F_\beta^w$ by $7.0\%$, $8.9\%$, and $10.9\%$ on the CHAMELEON \cite{skurowski2018animal_chameleon}, CAMO \cite{le2019anabranch_camo}, and COD10K \cite{Fan_2020_CVPR_sinet} dataset, respectively. Note that our method is also faster than SINet, \emph{i.e.}, 72 versus 51 FPS.

\begin{figure*}[tbp]
	\begin{center}
		\includegraphics[width = 1\linewidth]{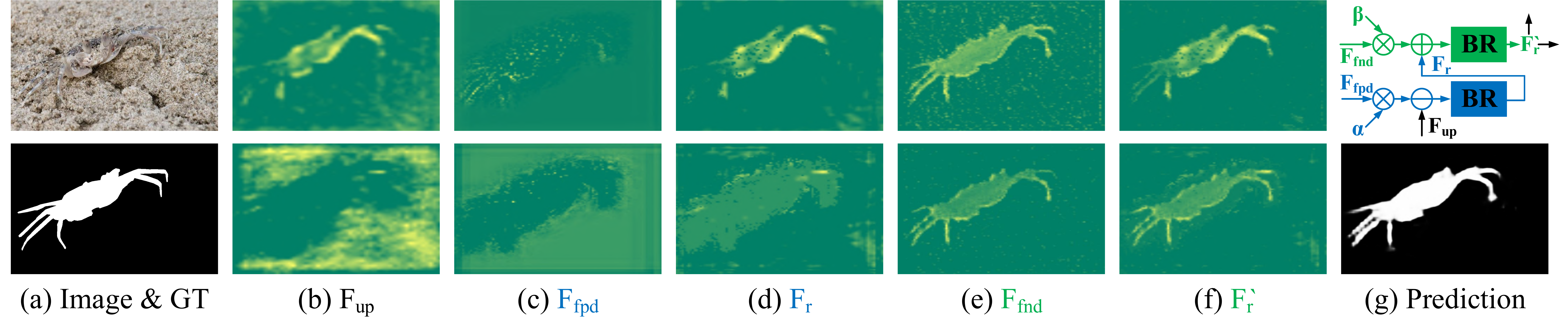}
	\end{center}
	\caption{Visualizing feature maps in the last FM. Best viewed in color and zoomed-in.}
	\label{fig:ablation}
\end{figure*}

Besides, Figure \ref{fig:visual} shows the qualitative comparison of our method with the others. It can be seen that our method is capable of accurately segmenting small camouflaged objects (\emph{e.g.}, the first two rows), large camouflaged objects (\emph{e.g.}, 3-\emph{rd} and 4-\emph{th} rows), and multiple camouflaged objects (\emph{e.g.}, 5-\emph{th} and 6-\emph{th} rows). This is mainly because that the positioning module (PM) can provide the initial location of the camouflaged objects with different scales for the following distraction mining, via exploring long-range semantic dependencies. While the state-of-the-arts are typically confused by the background which shares similar appearance with the camouflaged objects (\emph{e.g.}, 7-\emph{th} row) or the foreground region that cluttered in the background (\emph{e.g.}, 8-\emph{th} row), our method can successfully infer the true camouflaged region. This is mainly contributed by the proposed distraction mining strategy which could help suppress the false-positive distraction region and augment the false-negative distraction region. Furthermore, benefited by the multi-scale context exploration in the  distraction mining process, our method can capture detailed distraction information and thus has the ability to finely segment the camouflaged objects with complex structures (\emph{e.g.}, the last row).

\begin{table}[t]
	\centering
	\footnotesize
	\setlength{\tabcolsep}{6.5pt}
	\begin{tabular}{cl|cccc}
		\hline
		\multicolumn{2}{c|}{\multirow{2}{*}{Networks}} & \multicolumn{4}{c}{COD10K-Test (2,026 images)} \\
		\cline{3-6}
		& & $S_\alpha$$\uparrow$ & $E_\phi^{ad}$$\uparrow$ & $F_\beta^w$$\uparrow$ & $M$$\downarrow$ \\
		
		\hline
		(a) & B & 0.779 & 0.803 & 0.591 & 0.051 \\
		\hline
		(b) & B + CA & 0.788 & 0.819 & 0.618 & 0.046 \\
		(c) & B + SA & 0.791 & 0.826 & 0.624 & 0.046 \\
		(d) & B + PM & 0.792 & 0.835 & 0.631 & 0.045 \\
		\hline
		(e) & B + FPD & 0.790 & 0.844 & 0.632 & 0.043 \\
		(f) & B + FND & 0.790 & 0.837 & 0.628 & 0.043 \\
		(g) & B + FM \textit{w/o} A & 0.796 & 0.843 & 0.639 & 0.042 \\
		(h) & B + FM & 0.797 & 0.860 & 0.649 & 0.041 \\
		\hline
		(i) & B + PM + FPD & 0.796 & 0.854 & 0.645 & 0.042 \\
		(j) & B + PM + FND & 0.796 & 0.847 & 0.644 & 0.043 \\
		(k) & B + PM + FM \textit{w/o} A & 0.796 & 0.851 & 0.647 & 0.042 \\
		
		\hline
		(l) & PFNet & \textbf{0.800} & \textbf{0.868} & \textbf{0.660} & \textbf{0.040}     \\
		\hline
	\end{tabular}
	\caption{Ablation analyses. ``B'' denotes our network with the channel attention block (``CA'') and spatial attention block (``SA'') removed from positioning module (``PM'') and the false-positive distraction stream (``FPD'') and false-negative distraction stream (``FND'') in the focus module (``FM'') replaced by a simple skip-connection. ``\textit{w/o} A'' denotes that the higher-level prediction is not used as the attention map to guide the current-level features in the focus module. As can be observed, each proposed component plays an important role and contributes to the performance.}
	\label{tab:ablation}
\end{table}

\subsection{Ablation Study}
We conduct ablation studies to validate the effectiveness of two key components tailored for accurate camouflaged object segmentation, \emph{i.e.}, positioning module (PM) and focus module (FM), and report the results in Table \ref{tab:ablation}.

\textbf{The effectiveness of PM.} In Table \ref{tab:ablation}, we can see that adding the channel attention block (b) or the spatial attention block (c) on the base model (a) can boost the segmentation performance to some extent and the combination of the two (d) can achieve better results. This confirms that the PM can benefit the accurate camouflaged object segmentation.

\textbf{The effectiveness of FM.} Based on (a), introducing our proposed false-positive distraction mining (e) or false-negative distraction mining (f) would greatly improve the segmentation results. Considering both two types of distractions, \emph{i.e.}, (h), we obtain better results. For example, adding the focus module gains $5.7\%$ and $5.8\%$ performance improvement in terms of $E_\phi^{ad}$ and $F_\beta^w$, respectively. This shows that the FM enables our approach to possess the strong capability of accurately segmenting the camouflaged objects. When removing the guidance from the higher-level prediction, \emph{i.e.}, (g), the performance would decline to some extent. This is because that indiscriminately mining distractions from the input features increases the difficulty of the distraction discovery and thus hinder the effective distraction removal. This validates the rationality of our design to learn distractions from the attentive input features. 
From the results of (i-l), we can see that the above conclusions still keep true when adding the partial/full focus module on (d).
In addition, we visualize the feature maps in the last FM in Figure \ref{fig:ablation}. By mining the false-positive distractions (c), the false-positive predictions in (b) can be greatly suppressed (d). Through mining false-negative distractions (e), the purer representation of the target object can be obtained (f). This clearly demonstrates the effectiveness of the proposed distraction mining strategy which is designed to discover and remove distractions.

\section{Conclusion}
In this paper, we strive to embrace challenges towards accurate camouflaged object segmentation. We develop a novel distraction mining strategy for distraction discovery and removal. By adopting the distraction mining strategy in our bio-inspired framework, \emph{i.e.}, positioning and focus network (PFNet), we show that our approach achieves state-of-the-art performance on three benchmarks.
In the future, we plan to explore the potential of our method for other applications such as polyp segmentation and COVID-19 lung infection segmentation and further enhance its capability for segmenting camouflaged objects in videos.

\vspace{4mm}
\textbf{Acknowledgements.} This work was supported in part by the National Natural Science Foundation of China under Grant 91748104, Grant 61972067, Grant U1908214, in part by the Innovation Technology Funding of Dalian (Project No. 2020JJ26GX036).

\clearpage
{\small
\bibliographystyle{ieee_fullname}
\bibliography{paper}
}

\end{document}